%% file: emnlp2023.tex
\pdfoutput=1
\documentclass[11pt]{article}

\usepackage[]{EMNLP2023}

\usepackage{times}
\usepackage{latexsym}
\usepackage{listings}
\usepackage{multirow}
\usepackage[T1]{fontenc}
\usepackage[utf8]{inputenc}

\usepackage{microtype}

\usepackage{inconsolata}

\usepackage{graphicx}
\usepackage{booktabs}
\usepackage{amsmath}
\usepackage{gensymb}
\usepackage{mdframed}

\title{Text-to-OverpassQL: A Natural Language Interface for Complex Geodata Querying of OpenStreetMap}

\author{Michael Staniek\Thanks{equal contribution \enskip $^\diamondsuit$work done while at Computational Linguistics, Heidelberg University}\hspace{0.15cm}$^{1}$ \hspace{0.5cm} Raphael Schumann$^{*1}$ \hspace{0.4cm} Maike Züfle$^{\diamondsuit1,2}$ \hspace{0.4cm} Stefan Riezler$^{1,3}$ \\
  $^{1}$Computational Linguistics, Heidelberg University, Germany \\
  $^{2}$School of Informatics, University of Edinburgh, UK \\
  $^{3}$IWR, Heidelberg University, Germany \\
  \texttt{\{staniek,rschuman,zuefle,riezler\}@cl.uni-heidelberg.de}\\
  }

\begin{document}
\maketitle
\begin{abstract}
We present Text-to-OverpassQL, a task designed to facilitate a natural language interface for querying geodata from OpenStreetMap~(OSM). The Overpass Query Language~(OverpassQL) allows users to formulate complex database queries and is widely adopted in the OSM ecosystem. Generating Overpass queries from natural language input serves multiple use-cases. It enables novice users to utilize OverpassQL without prior knowledge, assists experienced users with crafting advanced queries, and enables tool-augmented large language models to access information stored in the OSM database. In order to assess the performance of current sequence generation models on this task, we propose OverpassNL\footnote{\url{https://github.com/raphael-sch/OverpassNL}}, a dataset of 8,352 queries with corresponding natural language inputs. We further introduce task specific evaluation metrics and ground the evaluation of the Text-to-OverpassQL task by executing the queries against the OSM database. We establish strong baselines by finetuning sequence-to-sequence models and adapting large language models with in-context examples. The detailed evaluation reveals strengths and weaknesses of the considered learning strategies, laying the foundations for further research into the Text-to-OverpassQL task.
\end{abstract}

\section{Introduction}
\begin{figure}[ht]
    \centering
    \includegraphics[width=0.48\textwidth]{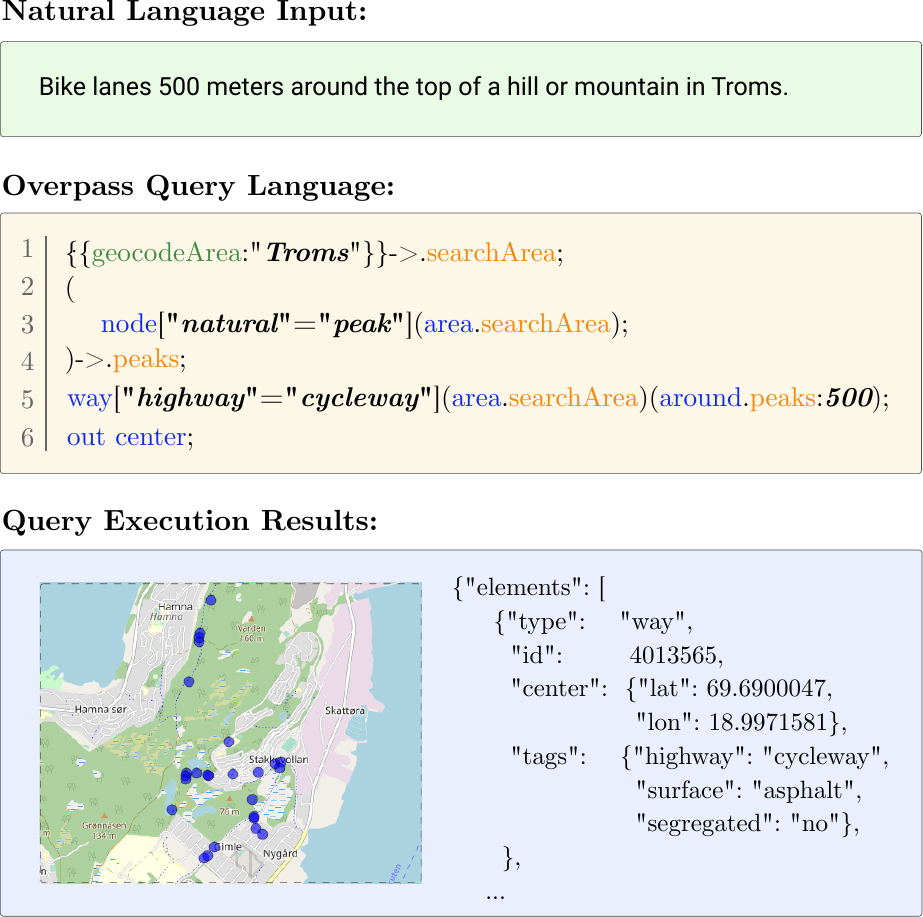}
    \caption{Natural language input and the corresponding Overpass query. The query is executed against the OpenStreetMap database and returns the requested elements in a structured response. The Overpass query language is highly expressive and allows to formulate complex queries to extract information from OpenStreetMap. Blue tokens in the query are syntax keywords, orange tokens are variable names and bold tokens define semantic properties of the requested elements. The green token in curly brackets geolocates an area called "Troms".}
    \label{fig:figure1}
\end{figure}

The OpenStreetMap~(OSM) database stores vast amounts of structured knowledge about our world. Users mainly access it via applications that render a visual map of inquired areas. A more advanced and systematic way to examine the stored information is to query the underlying geodata using the Overpass Query Language~(OverpassQL). OverpassQL is a feature-rich query language that is widely adopted in the OSM ecosystem, by contributors, analysts, and applications. In order to make this empowering query language accessible via natural language, we propose the Text-to-OverpassQL task. As is shown in Figure~\ref{fig:figure1}, the objective is to take a complex data request in natural language and translate it to OverpassQL in order to execute it against the OpenStreetMap database. Several groups of users can benefit from such a natural language interface. Inexperienced users are spared from learning the OverpassQL syntax. Expert users can use it to draft Overpass queries and then manually refine them, saving time and mental load in comparison to writing the full query from scratch. Another use-case would be to incorporate the interface as an API tool~\cite{schick2023toolformer} for large language models~(LLMs).

In this work, we introduce the three components that enable the Text-to-OverpassQL task. First, we present OverpassNL, a dataset of 8.5k natural language inputs and corresponding Overpass queries. The queries were collected from an OSM community website where they were written by OSM users to fulfill legitimate information needs. We then hired and trained students to write natural language descriptions of the queries. Second, we introduce a systematic evaluation protocol that assesses the prediction quality of a candidate system. To this end, we propose a task-specific metric that takes the similarity of the system output to Overpass queries on the levels of surface string, semantics, and syntax into account. Moreover, we ground the evaluation by executing the generated query against the OSM database and compare the returned elements with those returned by the gold query. Third, we explore several models and learning strategies to establish a base performance for the problem of generating Overpass queries from natural language. We finetuned sequence-to-sequence models and found that explicitly pretraining on code is not helpful for the task. We further explored in-context learning strategies for black-box LLMs and found that GPT-4~\cite{OpenAI2023GPT4TR} with few-shot examples, retrieved by sentence similarity, yields the best results, outperforming the finetuned sequence-to-sequence models. 

The proposed Text-to-OverpassQL task is a novel semantic parsing problem that is well motivated in real-world applications. While it shares characteristics with the Text-to-SQL task and its accompanying datasets, there are several key differences. The Text-to-OverpassQL task is grounded in a database that is genuinely in use and of global scale. The database is not divided into sub-databases or tables, and each Overpass query can retrieve any of the billions of stored elements in OSM. The desired elements have to be queried by a geographical specification and additional semantic tags. The tags are composed of key-value pairs that follow established community guidelines and conventions, but can also be open-vocabulary. Overall, the proposed task builds on a decades-long effort to structure and store the geographical world around us in a way to make it computationally accessible. With this work, we offer all components to benchmark future semantic parsing systems on a challenging real-world task.

Our main contributions are as follows: (i) We present OverpassNL, a dataset of 8.5k natural language inputs paired with real-world Overpass queries. (ii) We define task-specific evaluation metrics that take the OverpassQL syntax into account and are grounded in database execution. (iii) We train and evaluate several state-of-the-art sequence generation models to establish base performance and to identify specific properties of the proposed Text-to-OverpassQL task.

\section{Background}
\subsection{OpenStreetMap}
OpenStreetMap~(OSM) is a free and open geographic database that has been created and is maintained by a global community of voluntary contributors. The ever growing community has over 10M registered members who have collectively contributed to the creation of the existing 9B elements in the database~\cite{wiki:stats}. Elements are either nodes, ways, or relations. Nodes are annotated with geospatial coordinates. Ways are composed of multiple nodes and represent roads, building outlines, or area boundaries. Relations describe the relationships of elements, e.g., forming a municipal or major highway. Elements can be tagged with key-value pairs that assign semantic meaning and meta information. The OSM database is widely used in geodata analysis, scientific research, route planning applications, humanitarian aid projects, or augmented reality games. It also serves as a data source for geospatial services of companies like Facebook, Amazon or Apple~\cite{osm:consumers}.

\subsection{Overpass Query Language}
The Overpass Query Language~(OverpassQL) is a "procedural, imperative programming language written with a C style syntax"~\cite{wiki:oql}. It is used to query the OpenStreetMap database for geographic data and features. OverpassQL allows for detailed queries that are capable of extracting elements based on specific criteria, such as certain types of buildings, streets, or natural features within a defined area. Users can specify the types of elements they are interested in, and filter them by their associated key-value pairs.

\paragraph{Query Syntax}
We briefly explain the OverpassQL syntax based on the query depicted in Figure~\ref{fig:figure1} and refer to the official language guide for more information\footnote{\url{https://wiki.openstreetmap.org/wiki/Overpass_API/Language_Guide}}. The keyword \textit{geocodeArea} in the first line triggers a geolocation service\footnote{\url{https://nominatim.org/}} to find an area named "\textit{Troms}". The retrieved area is then assigned to a variable named \textit{searchArea}. The third line queries for nodes that are tagged with the \textit{natural=peak} key-value pair. The search is limited to nodes that are geographically within the previously defined \textit{searchArea}. The nodes that fullfill these criteria are stored into the \textit{peaks} variable. Line 5 queries for ways within the same area that are tagged with \textit{highway=cycleway} and are within a radius of 500 meters around a node stored in \textit{peaks}. Finally, the query requests a return of the specified ways.

\section{Related Work}
\paragraph{Natural Language Interfaces for Geodata}
One of the first attempts to build a natural language interface for geographical data is \textsc{GeoQuery}~\cite{zelle:aaai96, 10.5555/1619499.1619504}. It is a system based on Prolog, later adapted to SQL, and is tailored for a small database of U.S. geographical facts. Following works proposed methods to map the text input to the structured query language~\cite{10.5555/3020336.3020416, WongRaymond}. A more recent attempt is NLmaps~\cite{haas2016, lawrence2016} which aimed to build a natural language interface for OpenStreetMap. For querying the database, they designed a machine readable language~(MRL). The MRL is an abstraction of the Overpass Query Language, but it supports only a limited number of its features. To facilitate building more potent neural sequence-to-sequence parsers, they later released NLmaps v2~\cite{lawrence2018a} with augmented text and query pairs. Our work aims to support the Overpass Query Language without simplifications or abstractions, allowing to fully leverage the effort of the OpenStreetMap community that developed a query language that is optimally suited for the large-scale geospatial information in the OSM database.

\paragraph{Text-to-SQL}
The Text-to-SQL task~\cite{10.1007/3-540-44795-4_40, iyer-etal-2017-learning, 10.14778/2735461.2735468, wikisql} is closely related to the proposed Text-to-OverpassQL task and aims to provide a natural language interface to relational databases. While most of the works in this area focus on a specific domain and database, the Spider~\cite{yu-etal-2018-spider} dataset provides text and queries for multiple databases spanning different domains. They emphasize the hurdles of collecting real databases with complex schemas and sufficient data records, and circumvent this problem by mainly sourcing the databases from educational material and populate them with synthetic data. In contrast, the OpenStreetMap database is of global scale and is used in real-world applications. We highlight more differences between the datasets and underlying databases in Section~\ref{sec:comparison}. The Text-to-SQL task is commonly treated as a sequence-to-sequence problem~\cite{BRIDGE:Xi, yu2021grappa}. Some methods explicitly encode the database schema~\cite{zhang-etal-2019-editing}, while \citet{scholak-etal-2021-picard} show that finetuning a pretrained T5~model~\cite{Raffel2019ExploringTL} matches the performance of more specialized systems. They further introduced PICARD, a constraint decoding method that is SQL specific and enforces syntactic correctness. With the advent of large language models and in-context learning, more recent work focuses on prompt engineering the task~\cite{sun2023sqlpalm, chen2023teaching, pourreza2023dinsql}.

\begin{figure*}[ht]
    \centering
    \includegraphics[width=0.98\textwidth]{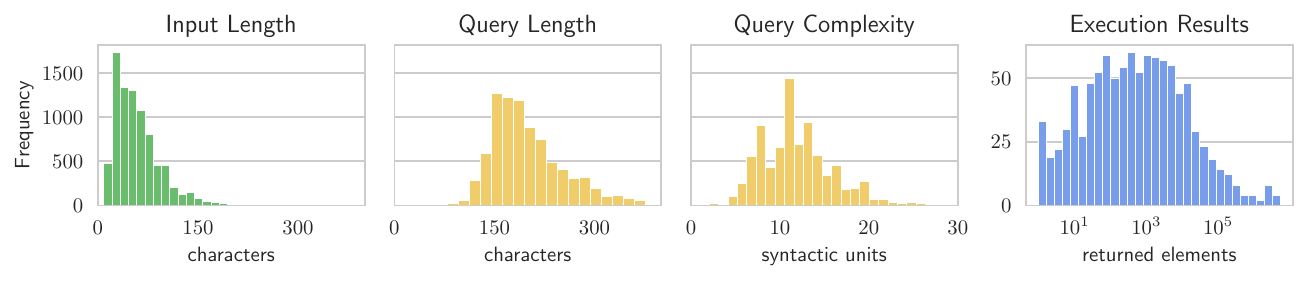}
    \caption{Dataset Statistics. Number of elements returned when executing the queries in the development set against OpenStreetMap. Each query returns at least one element and often several orders of magnitude more.}
    \label{fig:dataset_stats}
\end{figure*}
\begin{figure}[t]
    \centering
    \includegraphics[width=0.45\textwidth]{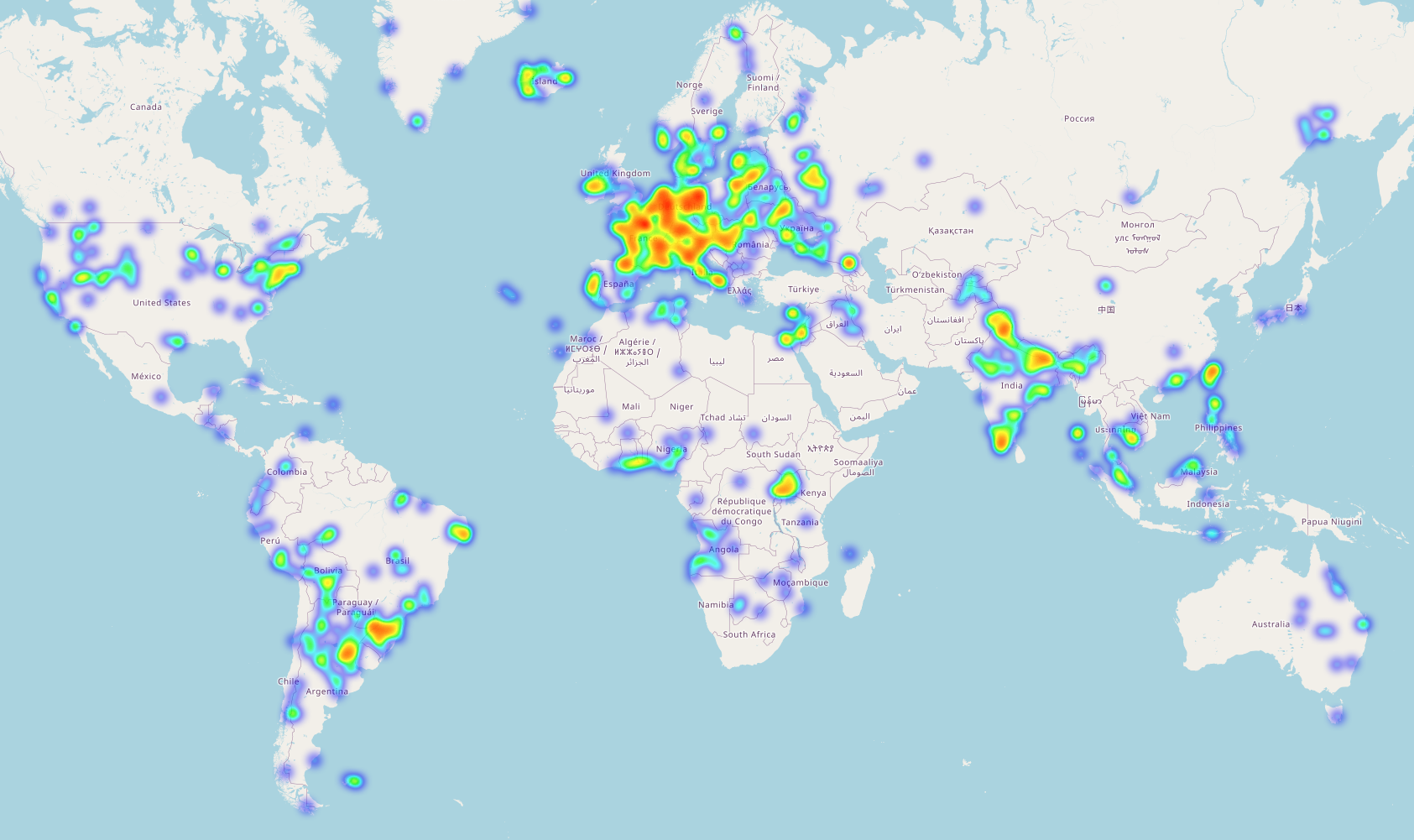}
    \caption{Location distribution of results returned by Overpass queries in our dataset. The queries cover locations on all continents. Europe is a traditional hotspot of the OpenStreetMap community and also has the best mapping coverage.}
    \label{fig:world_heatmap}
\end{figure}

\section{OverpassNL Dataset}
In order to facilitate the Text-to-OverpassQL task, we constructed a parallel dataset of natural language inputs and Overpass queries. This was done by collecting queries written and shared by users of Overpass Turbo\footnote{The shared queries constitute an unrestricted collection that is publicly available on \url{https://overpass-turbo.eu/}. The website is maintained by Martin Raifer, who helped us to acquire queries shared between 2014-2022.}, a web tool that allows to develop and execute Overpass queries within a graphical interface. We presented the queries to trained annotators who were tasked with writing natural language descriptions for them.

Initiating the dataset creation process with Overpass queries authored by real users and developers has several advantages: Firstly, the queries were created to satisfy legitimate information needs and, as such, cover a wide range of OverpassQL features. Also, the geographical coverage is high, as is shown in Figure~\ref{fig:world_heatmap} by the location distribution of elements returned by executing the queries in our dataset. Another advantage is that it is easier to teach annotators how to interpret Overpass queries than how to write them from scratch.

\subsection{Query Annotation}
For the annotation task, we recruited university students with proficient English skills and experience in database query languages like SQL. They had to complete a tutorial about OverpassQL and were subsequently tested on their knowledge. The test consisted of multiple choice questions and assignments to write text descriptions of pre-selected Overpass queries. Only the 15 students that passed this test were selected to participate in our annotation task. We built a graphical interface that showed them an Overpass query, the raw execution results, and the results rendered on a map. The task of the annotators was to write a natural language description that best represents the query. To ensure quality, we encouraged the annotators to use the Overpass documentation, continuously conducted spot tests on the submitted inputs, and required the annotators to validate the inputs written by other annotators. A screenshot of the annotation interface is shown in Appendix~\ref{apx:annotation}. We paid 100€ per 250 annotations, resulting in a wage of around 20€/hour.

\subsection{Dataset Statistics}
In total we obtained 8,352 queries annotated with natural language inputs. We split these into 6,352 instances for training, 1,000 for development, and 1,000 test instances. We constructed the splits such that there are no (near) duplicates on the input side or query side between training and evaluation instances. Figure~\ref{fig:dataset_stats} gives some statistics illustrating important dataset properties. There are a total of 11,259 distinct words in the natural language inputs, the mean input length is 59.7 characters, the mean query length is 199.8 characters, and each query has an average of 11.9 syntactic units. A syntactic unit is a subtree in the XML representation of an Overpass query. The rightmost plot in Figure~\ref{fig:dataset_stats} shows the number of elements returned by executing the development set queries.

\subsection{Complexity \& Coverage}
There are a total of 41 major syntax features in the Overpass Query Language. 31 of these 41 features occur in at least 20 queries of our dataset, resulting in a feature coverage rate of 76\%. See Appendix~\ref{apx:syntax_features} for a detailed list of syntax features. Our queries utilize 1,046 unique keys to specify tagged elements. These keys cover 91\% of all key usage in OpenStreetMap (see Appendix~\ref{apx:coverage}). The coverage of corresponding values is harder to estimate because of open-class keys like \textit{name}, \textit{source} or \textit{operator}. There are also keys that have recommended sets of values, e.g., the key \textit{leisure} is commonly paired with \textit{swimming\_pool}, \textit{skatepark} or \textit{pitch}. In total, we count 3,879 unique values and 4,880 unique key-value pairs in our dataset queries.

\subsection{Comparison to Other Datasets}
\label{sec:comparison}
\input{tables/comparison}
Because we are the first to present a dataset for the Text-to-OverpassQL task, we compare it to datasets of related tasks (see Table~\ref{tab:comparison}). \textsc{GeoQuery}~\cite{zelle:aaai96} is a small-scale dataset with 880 instances that allow to query 937 different geographical facts about the United States. NLmaps~\cite{haas2016} and NLmap~v2~\cite{lawrence2018a} provide queries for OpenStreetMap paired with natural language inputs, however, the queries are written in their own restricted query language called MRL. In contrast, we generate queries in the well-established OverpassQL language that has more features and is widely used by the OpenStreetMap community. Additionally, the NLmaps datasets only include up to 347 distinct keys in key-value pairs, limiting the semantic expressiveness of generated queries. Furthermore, NLmaps was built for an older version of the OSM database comprising one third of the size of the current version used in our work. WikiSQL~\cite{wikisql} converts single tables from Wikipedia articles to SQL databases and annotates queries with natural language inputs for them. While the dataset is of large scale, it contains only simple SQL queries and unconnected tables. The Spider dataset~\cite{yu-etal-2018-spider} includes 166 distinct databases of different domains and was collected for the Text-to-SQL task. Each natural language input in the dataset is intended for a specific database that is known a priori. This significantly reduces the number of relevant table names and column names per query, and simplifies the task by allowing to append known table and column names together with the database schema to the input. This stands in stark contrast to our Text-to-OverpassQL task where each query can utilize any key-value pair and retrieve elements from the entire OSM database, making it harder to predict the correct named identifiers for the desired elements. Also, the number of returned elements per query is orders of magnitude larger than in SQL related datasets. This makes the grounded evaluation harder by minimizing the likelihood of false positive matches in execution accuracy.

\section{Task \& Evaluation}
The Text-to-OverpassQL task requires to generate an Overpass query $\mathbf{q}$, given a natural language input $\mathbf{x}$. A model for this task aims to accurately translate the request, formulated in natural language, into code, written in the Overpass Query Language, that returns the correct elements when executed against the OpenStreetMap database. In order to evaluate such a system, we propose to use different evaluation metrics. These include metrics that compare the generated query with the reference query, and metrics that compare the results returned by executing the queries against the OSM database. In order to make the execution results reproducible, we release the evaluation script and a Docker container with the exact snapshot of the OSM database we used. We further describe the metrics in detail.

\subsection{Overpass Query Similarity Evaluation}
\label{sec:oqs}
We propose a language-specific metric called Overpass Query Similarity~(OQS) to quantify the compatibility of a generated query and the reference query. The metric is composed of three parts. First, we employ character F-score~(chrF) which measures the overlap of character n-grams between two strings~\cite{popovic-2015-chrf}. Because chrF operates on the character-level, it is well suited for Overpass queries which consist of words and special characters alike. Next, we calculate the overlap of keys and values between the generated query $\textbf{q}_G$ and reference query $\textbf{q}_R$. We define the Key Value Similarity~(KVS) as follows:
\begin{equation}
    \resizebox{.89\linewidth}{!}{KVS$(\textbf{q}_G, \textbf{q}_R) = \cfrac{| \text{KV}(\textbf{q}_G) \cap \text{KV}(\textbf{q}_R) |}{max(|\text{KV}(\textbf{q}_G)|, |\text{KV}(\textbf{q}_R)|)}$,}
    \label{equ:kvs}
\end{equation}
where the operator $\text{KV}(\cdot)$ returns all key-value pairs as well as all individual keys and all individual values. This metric captures the semantic relatedness of two queries. Complementary, the third part of the OQS metric compares queries on the syntactic level. We compute the Tree Similarity metric~(TreeS) by comparing the XML tree representation of the queries. We remove all key-value pairs and variable names from the trees and recursively compute the number of matching subtrees of the generated and reference query. Analogous to Equation~\ref{equ:kvs}, we normalize the number of matching subtrees by the maximum number of subtrees in either tree. Finally, the proposed Overpass Query Similarity metric is the mean of chrF, KVS and TreeS. The metric captures similarity of system outputs and reference queries on the levels of surface string, semantics, and syntax.

\subsection{Grounded Evaluation}
The nature of the Text-to-OverpassQL task allows us to perform a grounded evaluation of generated queries by executing them against the OpenStreetMap database. We quantify the correctness of the database execution by Execution Accuracy~(EX) which measures the exact match of all elements returned by executing the generated query and the reference query. Each returned element has an identifier number that is unique within OSM. We use this identifier number to determine exact matching of results. The plot on the right in Figure~\ref{fig:dataset_stats} shows that there are up to 10\textsuperscript{7} elements returned by a query. The matching by unique identifier and large number of returned elements make EX an inherently hard metric to satisfy.

Because OpenStreetMap is a community driven database that has grown over decades with changing annotation guidelines, there can be ambiguities in the tags of elements. For example, filtering all bridges can mean \textit{node["bridge"]} or \textit{node["bridge"="yes"]}. Both are correct according to current annotation guidelines, but do return slightly different sets of nodes. Another example is filtering for radar towers:
\begin{equation*}
\resizebox{\linewidth}{!}{\textit{node["man\_made"="tower"]["tower:type"="radar"].}}
\end{equation*}
The first filter is redundant according to tagging guidelines, but can not be omitted in queries because the guidelines are not consistently followed throughout the whole database. To account for this, we also report Soft Execution Accuracy~(EX\textsubscript{\textsc{soft}}). It is computed as the overlap of returned elements, normalized by the maximum number of elements returned by either the generated or reference query. The metric ranges from 0, which means no overlap, to 1, which is equivalent to exact match of results. We report the metric as percent in the results tables.

\section{Experiments}
\label{sec:exp}
In the following, we present experiments that showcase the opportunities to train machine learning models on the OverpassNL dataset and establish base performance of commonly used techniques. We finetuned sequence-to-sequence models of different sizes and pretraining settings. Additionally, we adapted black-box large language models with different in-context learning strategies. All models are evaluated with the proposed similarity and grounded metrics, as well as exact string match~(EM).

\subsection{Finetuning}
\input{tables/results_dev}
The task of generating Overpass queries from text is a sequence-to-sequence problem and can be addressed by a model with encoder-decoder architecture. The encoder processes the input text and the decoder autoregressively generates the Overpass query. In order to choose a suitable pretrained model, we focused on the T5 family of models because of their strong performance in a variety of sequence-to-sequence tasks~\cite{Raffel2019ExploringTL}, in particular Text-to-SQL~\cite{scholak-etal-2021-picard}. Besides the vanilla T5 model\footnote{The vanilla T5 model is not suitable for code or OverpassQL generation because it lacks tokens for '\{' or '[' in its vocabulary~\cite{wang2021codet5}.}, the family also includes models specifically trained for code~generation (CodeT5,~\citet{wang2021codet5}) and models with byte-level tokenization (ByT5,~\citet{xue-etal-2022-byt5}). Because neither model has been pretrained on data covering OverpassQL syntax, we ran experiments to compare the finetuning of CodeT5 and ByT5 on the training portion of our dataset. To expand the training set, we additionally created instances from comments that query authors put in some lines with the intention to describe the line's purpose and functionality. We extracted the comments and used them as the natural language input for the respective query. This produced 6,000 additional training instances.

The upper half of Table~\ref{tab:results_dev} shows the results for combinations of model type, model size, and training data, evaluated on the development set. In general, the \textit{base} variant of the models is better than the \textit{small} variant. We did not gain further improvements by finetuning even larger variants of the models. The results also show that the ByT5 models are better suited for our task than the CodeT5 models. Although the instances derived from developer comments are of low quality, they contribute to consistent improvements in execution accuracy for all models. The best model for our task is \textit{ByT5-base} with 582M parameters, finetuned on the enhanced training set. We further refer to this model as \textit{OverpassT5}.

\subsection{In-Context Learning}
\begin{figure}[t]
    \centering
    \includegraphics[width=0.48\textwidth]{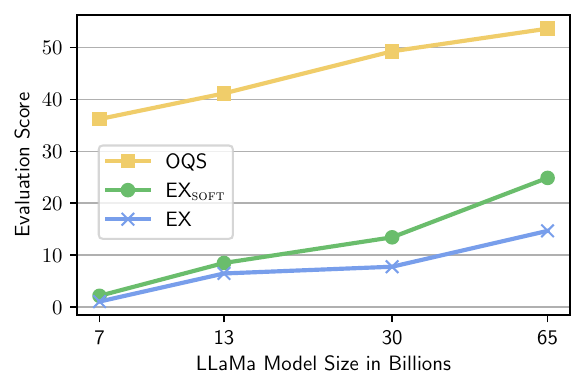}
    \caption{Development set results of LLaMa models with increasing number of parameters, prompted with five in-context examples.}
    \label{fig:model_size}
\end{figure}
\input{tables/results_test}
We furthermore explore the use of in-context learning for the Text-to-OverpassQL task. We prompt large language models to generate an Overpass query for the given text input while providing five example pairs as context. The structure of the 5-shot prompt is shown in Appendix~\ref{apx:prompt}. The example pairs are selected from the training set, either randomly or by input similarity~\cite{liu-etal-2022-makes}. We compare BLEU~\cite{papineni-etal-2002-bleu} and sentence-BERT embedding similarity~\cite{reimers-gurevych-2019-sentence} as metrics to retrieve the most similar examples.

Figure~\ref{fig:model_size} shows that the quality of queries generated by LLaMa~\cite{touvron2023llama} increases with the model size. The lower half of Table~\ref{tab:results_dev} shows results for the even bigger GPT-3~\cite{NEURIPS2020_1457c0d6} and GPT-4~\cite{OpenAI2023GPT4TR} models. We see a similar trend of improved results with increasing model size. Furthermore, retrieving similar instances from the training set as in-context examples is consistently better than a random selection. We also see that retrieval by sentence-BERT embedding similarity is better than retrieval by BLEU score. Best results are obtained for GPT-4 with sBERT retrieval, which will simply be referred to as \textit{GPT-4} in the following.

\subsection{Comparison between Finetuning and In-Context Learning}
In the previous sections, we selected the best models for finetuning and for in-context learning, based on development set performance. In order to compare the two models, we present results on the test set in Table~\ref{tab:results_test}~(top). While the surface metrics OQS and EM are nearly identical for both models, GPT-4 significantly outperforms OverpassT5 in the execution based metrics. It is also interesting that the queries generated by GPT-4 are more likely to raise syntax errors (\#Errors), despite achieving higher execution accuracy. Inspecting the individual components of OQS reveals that GPT-4 is better at generating correct key-value pairs, indicating that they are more important for correct execution results than faithfulness to the syntax of the reference query. We conjecture that the reason for this is that GPT-4 has likely seen a larger amount of OSM key-value pairs during pretraining than the finetuned models that are limited to OSM knowledge acquired from our training set. 

Table~\ref{tab:results_test}~(bottom) shows the results on the hard partition of the test set (defined in Section~\ref{sec:difficulty}). All performance metrics decrease significantly, without affecting the relative improvement of GPT-4 over OverpassT5. Notably, the majority of syntax errors stem from this partition of the test set. 

In sum, while GPT-4 is better at generating queries that return correct results, the OverpassT5 model is better at producing faithful OverpassQL syntax. However, GPT-4's advantage in execution accuracy comes at a computational cost. OverpassT5 is orders of magnitude smaller, resulting in faster inference speed and likely lower monetary cost per query. Also, GPT-4 is a 3rd-party API and does not allow for self-hosting, implying privacy concerns.

\section{Analysis}
\subsection{Instance Difficulty}
\label{sec:difficulty}
\begin{figure}[t]
    \centering
    \includegraphics[width=0.48\textwidth]{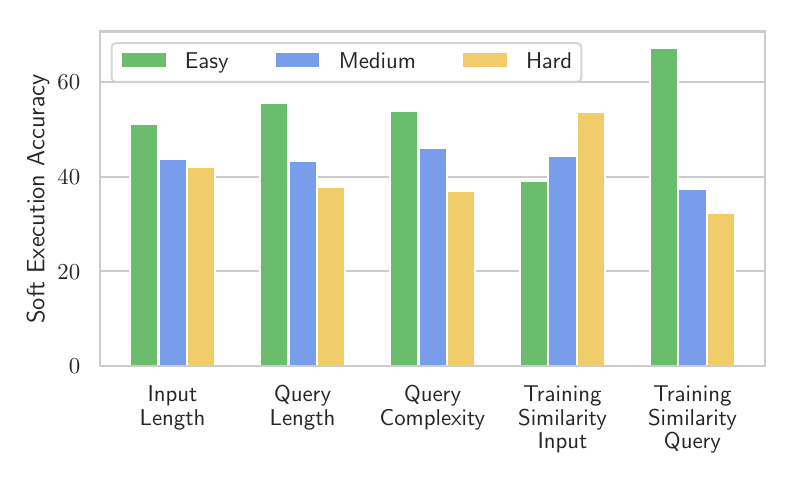}
    \caption{Instance difficulty on the development set using OverpassT5. Dividing the evaluation instances by highest similarity to any training query allows to measure performance on instances with different difficulties.}
    \label{fig:difficulty}
\end{figure}
To better assess the performance of our proposed models, we aim to divide the evaluation instances into three difficulty partitions. Figure~\ref{fig:difficulty} displays EX\textsubscript{\textsc{soft}} results for the easy, medium and hard partitions according to different difficulty criteria. A straightforward difficulty metric like the length of the input text has little influence on the accuracy. Using the length or complexity of the query (measured as the number of syntactic units) as difficulty metric leads to a clearer partition into easy, medium, and hard instances. Surprisingly, partitioning the instances based on their maximum input text similarity to any training instance leads to an undesired negative correlation of EX\textsubscript{\textsc{soft}} and difficulty where the instances with the lowest similarity to training inputs achieve best EX\textsubscript{\textsc{soft}} results. 
Finally, the clearest partition into instances of different difficulty is achieved by using the maximum similarity of a query to any query in the training set, where query similarity is measured by the OQS metric. A partitioning based on similarity to training queries can be seen as \textit{out-of-distribution} testing since it measures the performance for instances that are less likely to be memorized from the training set. We thus use this criterion to select instances comprising the hard partition in the experiments in Section~\ref{sec:exp}.

\subsection{Human Expert Evaluation}
\begin{figure}[t]
    \centering
    \includegraphics[width=0.48\textwidth]{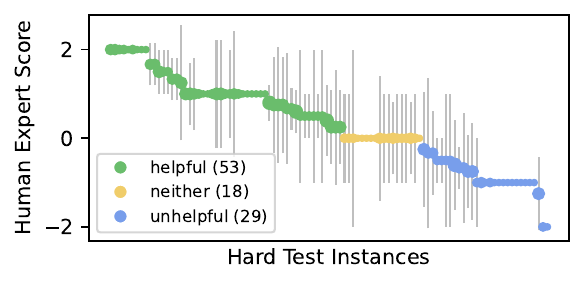}
    \caption{Human expert evaluation for 100 hard instances of the test set. Dot size indicates the number of human evaluators (up to 5 per instance). Grey bars depict the standard deviation.}
    \label{fig:survey}
\end{figure}
One motivation of the Text-to-OverpassQL task is to facilitate a system that assists expert users with crafting queries. While the surface and execution metrics allow us to compare different models, it is difficult to estimate how helpful imperfect queries are to human developers. To this end, we conducted an evaluation of the OverpassT5 outputs by human experts. They are Overpass developers that were recruited by postings in Overpass communities and they participated voluntarily in our experiment. They were asked to rate the helpfulness of a query given the input text on a five-point scale from "very unhelpful" to "very helpful". There were seven experts casting a total of 228 votes across 100 instances of the hard partition. In Appendix~\ref{apx:survey}, we explain the survey design in more detail. The results in Figure~\ref{fig:survey} show that the majority of generated queries were rated helpful, and less than a third were deemed unhelpful. This shows that even for the hardest test instances, the OverpassT5 model generates queries that are mostly helpful for developers when crafting a new query.

\subsection{Self-Refinement from Execution Feedback}
\input{tables/results_self_refine}
Recently, studies have shown that LLMs are able to self-refine their own outputs~\cite{madaan2023selfrefine}. The generated hypothesis is appended to the context and the LLM is prompted to generate an improved version. We conduct self-refine experiments for GPT-4 by appending the generated query and by additionally providing feedback from the query execution. If the query cannot be executed, we use the error message as feedback, otherwise we append a sample of the returned elements to the prompt. Table~\ref{tab:results_self_refine} shows results for self-refinement in two scenarios where either self-refinement is applied to all queries, or only to queries that raised a syntax error during execution. The results show that only refining syntax errors reduces the error count from 24 to 7 if an explicit error message is appended to the prompt, compared to a reduction to 19 errors if only the generated query is appended. However, this only leads to a slight increase in execution accuracy. On the other hand, refining all instances leads to an increase in errors, but also improves EX\textsubscript{\textsc{soft}} by 1.5 points when providing explicit feedback. These experiments show that there is still room for improving query generation with clever prompting techniques. Examples of feedback types and their effect on the generated queries can be found in Appendix~\ref{apx:refine_feedback}.

\section{Conclusion}
We introduced a novel semantic parsing task, called Text-to-OverpassQL. The objective of this task is to generate Overpass queries from natural language inputs. We highlighted its relevance within the OpenStreetMap ecosystem and related it to similar tasks. We identified key differences to the Text-to-SQL that pose unique challenges. To facilitate research on the task, we proposed OverpassNL, a dataset of real-world Overpass queries and corresponding annotations with natural language inputs. We used this dataset to train several state-of-the-art models and establish a base performance of the task. In order to measure prediction performance, we proposed task-specific metrics that take the OverpassQL syntax into account and are grounded in database execution. We presented a detailed evaluation of results that reveals the strengths and weaknesses of the considered learning strategies. We hope that our works serves as a foundation for further research on the challenging task of semantic parsing of geographical information grounded in the large and widely used OpenStreetMap database.

\section*{Acknowledgements}
The research reported in this paper was supported by a Google Focused Research Award on "Learning to  Negotiate Answers in Multi-Pass Semantic Parsing". We also thank Martin Raifer for helping us to acquire queries shared by Overpass Turbo users.

% Entries for the entire Anthology, followed by custom entries
\bibliography{anthology}
\bibliographystyle{acl_natbib}

\clearpage
\appendix
\input{appendix}

\end{document}

%% file: tables/comparison.tex
\begin{table*}[ht]
\centering
%\ra{1.05}
\resizebox{.99\linewidth}{!}{
\begin{tabular}{@{}l|cccccccccc@{}}
\toprule
\multirow{2}{*}{\textbf{Dataset}} & \multirow{2}{*}{\shortstack{\textbf{Query}\\\textbf{Language}}} & \multirow{2}{*}{\textbf{\shortstack{Number of\\Instances}}} & \multirow{2}{*}{\textbf{\shortstack{Query\\Templates}}} & \multicolumn{2}{c}{\textbf{Named Identifiers}} & \phantom{} & \multicolumn{2}{c}{\textbf{Extractable Elements}}& \phantom{}& \textbf{Results}\\
\cmidrule{5-6} \cmidrule{8-9} \cmidrule{11-11}
& & & & total & per database && total & per database && per query\\
\toprule
\textsc{GeoQuery} & SQL & 880 & 234 & 31 \& 7 & 31 \& 7 && 937 & 51 && 5\\
NLmaps & MRL & 2,380 & 379 & 107 & 107 && 3.4B & 3.4B && -\\
NLmaps v2 & MRL & 28,609 & 360 & 347 & 347 && 3.4B & 3.4B && -\\
WikiSQL & SQL & 81,654 & - & 168k & 6.38 && 460k & 11 && 1\\
Spider & SQL & 10,181 & 5,693 & 4,669 \& 876 & 58 \& 5 && 1.6M & 9.6k && 30\\\midrule
OverpassNL & OverpassQL & 8,352 & 3,890 & 1,046 & 1,046 && 9B & 9B && 10k\\
\bottomrule
\end{tabular}
}
\caption{Comparison of Text-to-Query datasets. Templates are normalized queries, i.e. removing named identifiers, variable names and digits. Named identifiers are tag keys in OpenStreetMap related datasets and table \& column names in SQL related datasets. The Spider dataset includes 166 unconnected databases and the task is to generate a query for a specific database that is known a priori. Thus, the average number of relevant named identifiers and extractable elements is much smaller per query than for the whole dataset.}

\label{tab:comparison}
\end{table*}

%% file: tables/results_dev.tex
\begin{table*}[t]
\centering
%\ra{1.05}
\resizebox{.99\linewidth}{!}{
\begin{tabular}{l@{}llccccccccc}
\toprule
&\multicolumn{1}{c}{\textbf{}}&&  \multicolumn{4}{c}{\textbf{Overpass Query Similarity}} && &&\multicolumn{2}{c}{\textbf{Execution Accuracy}}\\ 
\cmidrule{4-7} \cmidrule{11-12}
    \textbf{Model} & \multicolumn{1}{l}{\textbf{Setting}} &\phantom{} & chrF & KVS & TreeS & \textbf{OQS}& \phantom{} & \textbf{EM}& \phantom{} & \textbf{EX} & \textbf{EX\textsubscript{\textsc{soft}}}\\
    \toprule
\midrule
    \multicolumn{12}{c}{\textbf{Finetuning}}\\
\midrule
CodeT5-small     &   && 74.0 \scriptsize{±0.2} & 61.2 \scriptsize{±0.6} & 72.2 \scriptsize{±0.1} & 69.1 \scriptsize{±0.2} && 18.5 \scriptsize{±0.1} && 31.9 \scriptsize{±0.9} & 41.9 \scriptsize{±0.6}\\
CodeT5-small    & \enskip +comments  && 74.1 \scriptsize{±0.2} & 62.4 \scriptsize{±0.7} & 72.7 \scriptsize{±0.3} & 69.8 \scriptsize{±0.2} && 18.9 \scriptsize{±0.4} && 32.7 \scriptsize{±0.6} & 43.9 \scriptsize{±0.5}\\
CodeT5-base     &   && 74.6 \scriptsize{±0.0} & 63.2 \scriptsize{±0.4} & 73.0 \scriptsize{±0.1} & 70.3 \scriptsize{±0.2} && 19.8 \scriptsize{±0.4} && 33.3 \scriptsize{±0.3} & 44.3 \scriptsize{±0.1}\\
CodeT5-base    & \enskip +comments  && 74.9 \scriptsize{±0.1} & 63.6 \scriptsize{±0.5} & 73.5 \scriptsize{±0.1} & 70.7 \scriptsize{±0.2} && 20.3 \scriptsize{±0.2} && 34.5 \scriptsize{±0.4} & 46.2 \scriptsize{±0.4}\\
\midrule
ByT5-small    &   && 74.8 \scriptsize{±0.2} & 64.4 \scriptsize{±0.2} & 73.1 \scriptsize{±0.2} & 70.8 \scriptsize{±0.2} && 20.4 \scriptsize{±0.1} && 35.4 \scriptsize{±0.3} & 46.8 \scriptsize{±0.3}\\
ByT5-small    & \enskip +comments  && 75.0 \scriptsize{±0.0} & 64.6 \scriptsize{±0.2} & 73.4 \scriptsize{±0.2} & 71.0 \scriptsize{±0.1} && 21.0 \scriptsize{±0.3} && 36.0 \scriptsize{±0.4} & 46.2 \scriptsize{±0.5}\\
ByT5-base     &   && \textbf{75.5} \scriptsize{±0.0} & 65.0 \scriptsize{±0.2} & \textbf{73.8} \scriptsize{±0.2} & 71.4 \scriptsize{±0.0} && 21.9 \scriptsize{±0.4} && 36.2 \scriptsize{±0.0} & 46.7 \scriptsize{±0.4}\\
ByT5-base    & \enskip +comments  && \textbf{75.5} \scriptsize{±0.1} & \textbf{66.0} \scriptsize{±0.2} & 73.7 \scriptsize{±0.4} & \textbf{71.7} \scriptsize{±0.1} && \textbf{22.0} \scriptsize{±0.1} && \textbf{36.7} \scriptsize{±0.6} & \textbf{47.0} \scriptsize{±1.0}\\
\midrule
\midrule
    \multicolumn{12}{c}{\textbf{5-Shot In-Context Learning}}\\
\midrule
GPT-3 & \enskip random && 58.8 \phantom{\scriptsize{±0.0}} & 48.8 \phantom{\scriptsize{±0.0}} & 57.5 \phantom{\scriptsize{±0.0}} & 55.0 \phantom{\scriptsize{±0.0}} && 4.1 \phantom{\scriptsize{±0.0}} && 16.9 \phantom{\scriptsize{±0.0}} & 28.0 \phantom{\scriptsize{±0.0}}\\
GPT-3 & \enskip retrieval-BLEU && 67.4 \phantom{\scriptsize{±0.0}} & 55.5 \phantom{\scriptsize{±0.0}} & 66.8 \phantom{\scriptsize{±0.0}} & 63.3 \phantom{\scriptsize{±0.0}} && 18.0 \phantom{\scriptsize{±0.0}} && 28.7 \phantom{\scriptsize{±0.0}} & 37.7 \phantom{\scriptsize{±0.0}}\\
GPT-3 & \enskip retrieval-sBERT && 72.1 \phantom{\scriptsize{±0.0}} & 63.3 \phantom{\scriptsize{±0.0}} & 69.9 \phantom{\scriptsize{±0.0}} & 68.4 \phantom{\scriptsize{±0.0}} && 19.5 \phantom{\scriptsize{±0.0}} && 34.1 \phantom{\scriptsize{±0.0}} & 44.2 \phantom{\scriptsize{±0.0}}\\
\midrule
GPT-4 & \enskip random && 63.8 \phantom{\scriptsize{±0.0}} & 57.5 \phantom{\scriptsize{±0.0}} & 61.1 \phantom{\scriptsize{±0.0}} & 60.8 \phantom{\scriptsize{±0.0}} && 5.1 \phantom{\scriptsize{±0.0}} && 25.4 \phantom{\scriptsize{±0.0}} & 39.5 \phantom{\scriptsize{±0.0}}\\
GPT-4 & \enskip retrieval-BLEU && 74.3 \phantom{\scriptsize{±0.0}} & 66.1 \phantom{\scriptsize{±0.0}} & 72.4 \phantom{\scriptsize{±0.0}} & 71.0 \phantom{\scriptsize{±0.0}} && 22.9 \phantom{\scriptsize{±0.0}} && 38.5 \phantom{\scriptsize{±0.0}} & 50.7 \phantom{\scriptsize{±0.0}}\\
GPT-4 & \enskip retrieval-sBERT && \textbf{75.7} \phantom{\scriptsize{±0.0}} & \textbf{69.9} \phantom{\scriptsize{±0.0}} & \textbf{74.0} \phantom{\scriptsize{±0.0}} & \textbf{73.2} \phantom{\scriptsize{±0.0}} && \textbf{23.4} \phantom{\scriptsize{±0.0}} && \textbf{40.4} \phantom{\scriptsize{±0.0}} & \textbf{53.0} \phantom{\scriptsize{±0.0}}\\
\bottomrule
\end{tabular}
}
\caption{Results on the \textbf{development set} of OverpassNL. The proposed Overpass Query Similarity~(OQS) metric is the mean of Character F-score~(chrF), Key-Value similarity~(KVS) and XML-tree similarity~(TreeS). EM denotes exact string match. Execution accuracy (EX) is the exact match of all results returned by executing the generated query and reference query against OpenStreetMap. Soft Execution Accuracy~(EX\textsubscript{\textsc{soft}}) is the normalized overlap of returned results. Results in \textbf{bold} are best for the respective learning setup. Finetuning experiments are repeated three times with different random seeds and mean/standard deviation are reported.}
\label{tab:results_dev}
\end{table*}

%% file: tables/results_test.tex
\begin{table*}[ht]
\centering
%\ra{1.05}
\resizebox{.95\linewidth}{!}{
\begin{tabular}{l@{}lccccccccccc}
\toprule
&\multicolumn{1}{c}{\textbf{}}&  \multicolumn{4}{c}{\textbf{Overpass Query Similarity}} && && &&\multicolumn{2}{c}{\textbf{Execution Accuracy}}\\ 
\cmidrule{3-6} \cmidrule{12-13}
    \textbf{Model} & \phantom{} & chrF & KVS & TreeS & \textbf{OQS}& \phantom{} & \textbf{EM}& \phantom{}& \textbf{\#Errors}& \phantom{} & \textbf{EX} & \textbf{EX\textsubscript{\textsc{soft}}}\\
    \toprule
\midrule
    \multicolumn{13}{c}{\textbf{Full Test Set (1,000 Instances)}}\\
\midrule
OverpassT5 && \textbf{74.9} \scriptsize{±0.1} & 66.1 \scriptsize{±0.3} & \textbf{72.7} \scriptsize{±0.2} & 71.2 \scriptsize{±0.2} && \textbf{20.7} \scriptsize{±0.2} && \textbf{23} \scriptsize{±5.6} && 33.9 \scriptsize{±0.1} & 46.3 \scriptsize{±0.3}\\

GPT-4 && 73.6 \phantom{\scriptsize{±0.0}} & \textbf{68.6} \phantom{\scriptsize{±0.0}} & 72.0 \phantom{\scriptsize{±0.0}} & \textbf{71.4} \phantom{\scriptsize{±0.0}} && \textbf{20.7} \phantom{\scriptsize{±0.0}} && 34 \phantom{\scriptsize{±0.0}} && \textbf{38.9} \phantom{\scriptsize{±0.0}} & \textbf{53.0} \phantom{\scriptsize{±0.0}}\\
\midrule
    \multicolumn{13}{c}{\textbf{Hard Partition (333 Instances)}}\\
\midrule
OverpassT5 && \underline{62.8} \scriptsize{±0.4} & 57.7 \scriptsize{±0.4} & \underline{56.6} \scriptsize{±0.3} & \underline{59.1} \scriptsize{±0.3} && 8.8 \scriptsize{±0.3} && \underline{15} \scriptsize{±4.5} && 18.7 \scriptsize{±0.1} & 29.7 \scriptsize{±0.5}\\
GPT-4 && 61.4 \phantom{\scriptsize{±0.0}} & \underline{59.6} \phantom{\scriptsize{±0.0}} & 56.3 \phantom{\scriptsize{±0.0}} & \underline{59.1} \phantom{\scriptsize{±0.0}} && \underline{9.0} \phantom{\scriptsize{±0.0}} && 25 \phantom{\scriptsize{±0.0}} && \underline{22.2} \phantom{\scriptsize{±0.0}} & \underline{35.3} \phantom{\scriptsize{±0.0}}\\
\bottomrule
\end{tabular}
}
\caption{Results on the \textbf{test set} of our OverpassNL dataset. The proposed Overpass Query Similarity~(OQS) metric is the mean of Character F-score~(chrF), Key-Value Similarity~(KVS) and XML-tree Similarity~(TreeS). Exact match (EM) is query string match. \#Errors are number of raised syntax errors when trying to execute the query. Execution accuracy (EX) is the exact match of all results returned by executing the model generated query and reference query against OpenStreetMap. Soft Execution Accuracy~(EX\textsubscript{\textsc{soft}}) is the normalized overlap of returned results. \textbf{Bold} results are best on the test set and \underline{underlined} results are best on the hard partition.}
\label{tab:results_test}
\end{table*}

%% file: tables/results_self_refine.tex
\begin{table}[t]
\centering
%\ra{1.05}
\resizebox{.99\linewidth}{!}{
\begin{tabular}{@{}llcccc@{}}
\toprule
    \textbf{Model} & \textbf{OQS} & \textbf{EM} & \textbf{\#Errors} & \textbf{EX} & \textbf{EX\textsubscript{\textsc{soft}}}\\
\midrule
GPT-4      &   \textbf{73.2} & \textbf{23.4} & 24 & 40.4 & 53.0\\\midrule
\multicolumn{6}{c}{\textbf{Refine Syntax Errors Only}}\\\midrule
\;no feedback      &   \textbf{73.2} & \textbf{23.4} & 19 & 40.4 & 53.1\\
\;with feedback   &   \textbf{73.2} & \textbf{23.4} & \textbf{7} & 40.7 & 53.7\\\midrule
\multicolumn{6}{c}{\textbf{Refine All Instances}}\\\midrule
\;no feedback      &   73.1 & 21.9 & 31 & 39.6 & 52.9\\
\;with feedback   &   73.1 & \textbf{23.4} & 26 & \textbf{41.4} & \textbf{54.5}\\
\bottomrule
\end{tabular}
}
\caption{Self-Refinement of hypotheses generated by GPT-4 on the \textbf{development set}. Feedback is either the error message if raised during execution or the returned results.}
\vspace{-0.18cm}
\label{tab:results_self_refine}
\end{table}

%% file: appendix.tex
\section{Key Usage}
\label{apx:coverage}
\begin{figure}[t]
    \centering
    \includegraphics[width=0.48\textwidth]{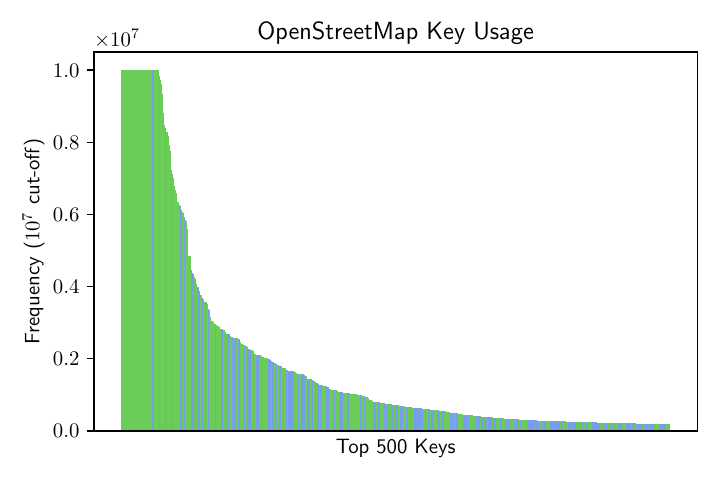}
    \caption{Top 500 keys by usage in OpenStreetMap. Green indicates that the key is used in our dataset and blue that it is not. The keys in our dataset cover 91\% of keys tagged across the entire OSM database.}
    \label{fig:key_usage}
\end{figure}

There are 90k unique keys in OpenStreetMap of which only 6k are actively used\footnote{\url{https://taginfo.openstreetmap.org/keys}}. We define active usage as 1k or more elements tagged with that key. Keys that are used less than 1k are often include typos, are outdated or leaked from proprietary tagging schemes. Figure~\ref{fig:key_usage} shows that the 1k unique keys in our dataset account for 91\% of all key usage in OpenStreetMap. Key Usage is the total number of times any key is used to tag an element.

\section{Model Details}
In the finetuning experiments, we use CodeT5\footnote{\url{https://huggingface.co/Salesforce/codet5-base}} and ByT5\footnote{\url{https://huggingface.co/google/byt5-base}} provided by huggingface. We finetune all weights (full finetuning) for 30 epochs using the Adam~\cite{adam} optimizer with weight decay of 0.1. The maximum learning rate is $4xs10^{-4}$ with warmup for 10\% of the steps and a linear decay schedule. Training batch size is 16 and we decode with four search beams. For in-context learning, we use the official OpenAI API to access GPT-3 (\textit{text-davinci-003}) and GPT-4 (\textit{gpt-4-0314}).

\section{Syntax Features}
\label{apx:syntax_features}
We count the occurrences of major OverpassQL syntax features used in the queries of our dataset. We use the OverpassQL article\footnote{\url{https://wiki.openstreetmap.org/w/index.php?title=Overpass_API/Overpass_QL&oldid=2568570}} in the OpenStreetMap Wiki to categorize features and list their prevalence in our dataset in Figure~\ref{table:features}. In total 31 of the 41 syntax features occur in at least 20 queries of our dataset. This represents a coverage of 76\%. The "Evaluators" category is a subcategory of "Conditional Query Filter" and we do not count each evaluator type as major feature.

\begin{figure}[t]
    \centering
    \includegraphics[width=0.48\textwidth]{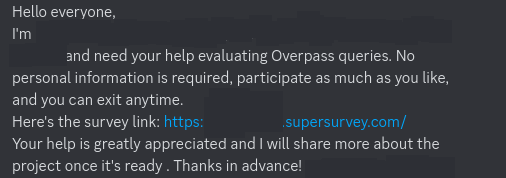}
    \caption{Post to announce our survey on the OpenStreetMap discord.}
    \label{fig:survey_announce}
\end{figure}

\input{tables/features}

\section{Prompts}
\label{apx:prompt}
The structure of the 5-shot prompt is shown in Figure~\ref{prompt:5shot} and the prompt for self refinement is shown in Figure~\ref{prompt:5shot_refine_feedback}. 

\input{prompts/5shot_other}
\input{prompts/5shot_refine_feedback_other}

\section{Refine+Feedback}
\label{apx:refine_feedback}
Examples of different feedback types (Empty Output, Error Message or Normal Output) and the effect they have on the generated query can be found in Figure \ref{tables:refine_examples}.
\input{tables/refine_feedback_examples}

\section{Human Expert Survey}
\label{apx:survey}
\begin{figure*}[t]
    \centering
    \includegraphics[width=0.98\textwidth]{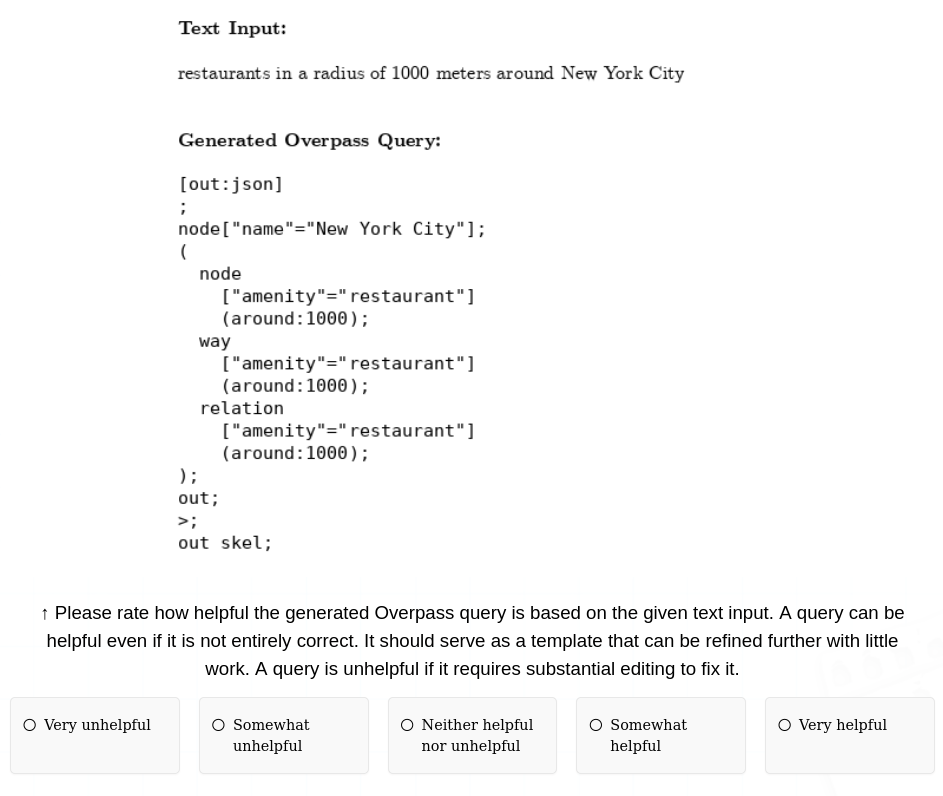}
    \caption{Interface of the expert survey. At the bottom are the answer options and our definition of helpfulness.}
    \label{fig:survey_item}
\end{figure*}

The link to the survey was posted on the Discord community called "Openstreetmap World" and the official OpenStreetMap Slack server. Both have a dedicated "Overpass" channel that we posted to. Figure~\ref{fig:survey_announce} shows the survey announcement and Figure~\ref{fig:survey_item} shows the survey interface including the answer options and our definition of helpfulness.

\section{Annotation Interface}
\label{apx:annotation}
We show the annotation interface in Figure~\ref{apx:annotation}. The annotators see the query on the top. The results of the query are visualized in the middle and the raw outputs are below. The text box on the bottom is used by the annotators to write the natural language input for the given query. 

\begin{figure*}[t]
    \centering
    \includegraphics[width=0.70\textwidth]{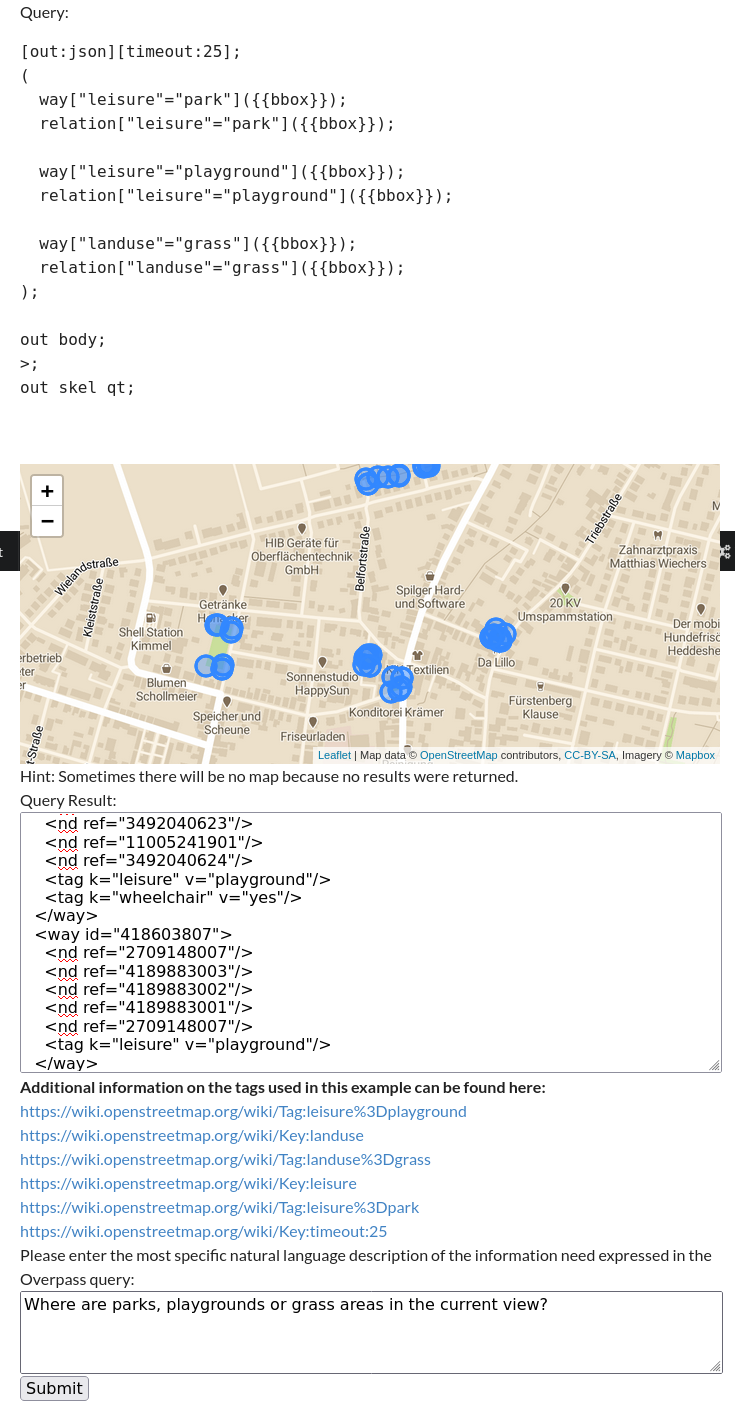}
    \caption{Annotation interface to collect the natural language descriptions for the queries in our dataset.}
    \label{fig:annotation}
\end{figure*}

%% file: tables/features.tex
\begin{figure}[ht]
\begin{mdframed}
\begin{itemize}
\small
    \item Settings (5/6)
    \begin{itemize}
        \item Timeout (5,985; 71.7\%)
        \item Element Limit (125; 1.5\%)
        \item Output Format (6,945; 83.2\%)
        \item Bounding Box (2,746; 32.9\%)
        \item Date (286; 3.4\%)
        \item Diff/adiff two dates (0; 0\%)
    \end{itemize}
    \item Block Statements (4/8)
    \begin{itemize}
        \item Union (7,651; 91.6\%)
        \item Difference (231; 2.8\%)
        \item if (5; 0.1\%)
        \item for-each (0; 0\%)
        \item for (85; 1.0\%)
        \item complete (20; 0.2\%)
        \item retro (1; 0.0\%)
        \item compare (0; 0\%)
    \end{itemize}
    \item Standalone Statements (9/13)
    \begin{itemize}
        \item out (8,186; 98.0\%)
        \item Item (1,115; 13.4\%)
        \item Recurse Up (30; 0.4\%)
        \item Recurse Up Relations (6; 0.1\%)
        \item Recurse Down (6,084; 72.8\%)
        \item Recurse Down Relations (54; 0.6\%)
        \item is\_in (22; 0.3\%)
        \item timeline (0; 0\%)
        \item local (0; 0\%)
        \item convert (8; 0.1\%)
        \item make (115; 1.4\%)
        \item The Query Statement (8,313; 99.5\%)
        \item The Query Filter (120; 1.4\%)
    \end{itemize}
    \item Filters (12/14)
    \begin{itemize}
        \item By Tag (7,860; 94.1\%)
        \item Bounding Box (2,548; 30.5\%)
        \item Recurse By nwr(402; 4.8\%)
        \item Recurse By Way Count (0; 0\%)
        \item By Input Set (8,206; 98.3\%)
        \item By Element Id (3,571; 42.8\%)
        \item Relative to Other Elements (389; 4.7\%)
        \item By Polygon (8; 0.1\%)
        \item Newer (286; 3.4\%)
        \item By Date of Change (345; 4.1\%)
        \item By User (611; 7.3\%)
        \item By Area (4,846; 58.0\%)
        \item Area Pivot (36; 0.4\%)
        \item Conditional Query Filter (236; 2.8\%)
    \end{itemize}
\end{itemize}
\end{mdframed}
\caption{List of Overpass Query Language syntax features and the prevalence in our dataset.}
\label{table:features}
\end{figure}

%% file: prompts/5shot_other.tex
\begin{figure*}[ht]
\begin{mdframed}%[frametitle={Prompt:}]
\small
The OverpassQL language allows one to formulate questions to the OpenStreetMap database.\\
\\
Here are a few examples:\\
\\
Input:\\
All historic castles in Germany.\\
\\
Overpass Query:
\vspace{-0.17cm}
\begin{lstlisting}
[out:xml][timeout:500];area["name"="Deutschland"]["admin_level"]->.a;
(node["historic"="castle"](area.a);way["historic"="castle"](area.a);
relation["historic"="castle"](area.a););
\end{lstlisting}
%<@\textbf{bold text}@>

\vspace{0.5cm}
Input:\\
Find every castle in Luxemburg,Neatherlands and Belgium.\\

Overpass Query:
\vspace{-0.17cm}
\begin{lstlisting}
[out:json][timeout:120];(({{geocodeArea:"Belgium"}}->.be;{{geocodeArea:"Luxembourg"}}->.lu;
{{geocodeArea:"Nederland"}}->.nl;)->.benelux;
(node["historic"="castle"]["name"](area.benelux);););out center;
\end{lstlisting}

\vspace{0.5cm}
Input:\\
Castles in current view.\\

Overpass Query:
\vspace{-0.17cm}
\begin{lstlisting}
[out:json][timeout:25];(node["historic"="castle"]({{bbox}});way["historic"="castle"]({{bbox}});
relation["historic"="castle"]({{bbox}}););out;>;out skel qt;
\end{lstlisting}

\vspace{0.5cm}
Input:\\
Castles in current view.\\

Overpass Query:
\vspace{-0.17cm}
\begin{lstlisting}
[out:json][timeout:25];(node["historic"="castle"]({{bbox}});
way["historic"="castle"]({{bbox}});relation["historic"="castle"]({{bbox}}););
out;>;out skel qt;
\end{lstlisting}

\vspace{0.5cm}
Input:\\
castles in Tuscany.\\

Overpass Query:
\vspace{-0.17cm}
\begin{lstlisting}
[out:json][timeout:250];{{geocodeArea:"Tuscany"}}->.searchArea;
(node["historic"="castle"](area.searchArea);way["historic"="castle"](area.searchArea);
relation["historic"="castle"](area.searchArea););out;>;out skel qt;
\end{lstlisting}

\vspace{0.5cm}
Input:\\
\textit{castle in Deutschland}\\

Overpass Query:

\textless \textgreater
\end{mdframed}
\caption{The \textbf{5-shot prompt} used to generate Overpass queries from natural language input. The in-context examples are selected from our training set by similarity to the current natural language input. We use sBERT to embed the natural language input and compute the cosine similarities. The current natural language input is in \textit{cursive} and \textless \textgreater~indicates the position where the LLM starts to generate next tokens.}
\label{prompt:5shot}
\end{figure*}

%% file: prompts/5shot_refine_feedback_other.tex
\begin{figure*}[ht]
\begin{mdframed}
\small
The OverpassQL language allows one to formulate questions to the OpenStreetMap database.
Your goal is, given an Input and a Hypothesis, to produce a improved version of the Hypothesis.
If the Hypothesis is already good enough, do not try to improve it.\\

Here are a few examples:\\

Input:\\
atms in Germany\\

Hypothesis:
\vspace{-0.17cm}
\begin{lstlisting}
[out:json][timeout:25];area["name"="Germany"]->.a;(node["amenity"="atm"](area.a);
way["amenity"="atm"](area.a);relation["amenity"="atm"](area.a););
out;>;out skel qt;
\end{lstlisting}

Overpass Query:
\vspace{-0.17cm}
\begin{lstlisting}
[out:json][timeout:25];{{geocodeArea:"Deutschland"}}->.searchArea;
(node["amenity"="atm"](area.searchArea);way["amenity"="atm"](area.searchArea);
relation["amenity"="atm"](area.searchArea););out center;
\end{lstlisting}

\vspace{0.5cm}
Input:\\
ATMs and banks with ATMs in Berlin.\\

Hypothesis:
\vspace{-0.17cm}
\begin{lstlisting}
[out:json][timeout:25];{{geocodeArea:"Berlin"}}->.searchArea;
(node["amenity"="atm"](area.searchArea);node["amenity"="bank"]["atm=yes"](area.searchArea););
out;out;>;out skel qt;
\end{lstlisting}

Overpass Query:
\vspace{-0.17cm}
\begin{lstlisting}
[out:json][timeout:25];
area["name"="Berlin"]->.a;(node["amenity"="bank"]["atm"="yes"](area.a);
node["amenity"="atm"](area.a);way["amenity"="bank"]["atm"="yes"](area.a);>;
way["amenity"="atm"](area.a);>;);out;
\end{lstlisting}

\begin{mdframed}
\centering
\textit{... three more in-context examples ...}
\end{mdframed}

\quad\\
Here is an Input:\\
Banks or ATMS in Bürggen of Kreis Viersen.\\

Here is the Overpass Query Hypothesis produced by a model:\\
\vspace{-0.17cm}
\begin{lstlisting}
[out:json][timeout:25]; {{geocodeArea:"Bürggen"}}->.searchArea;
{{geocodeArea:"Kreis Viersen"}}->.searchArea2;
(node["amenity"="bank"](area.searchArea)(area.searchArea2);
node["amenity"="atm"](area.searchArea)(area.searchArea2);
way["amenity"="bank"](area.searchArea)(area.searchArea2);
relation["amenity"="bank"](area.searchArea)(area.searchArea2); ); out; >; out skel qt;
\end{lstlisting}

You will now get part of the Overpass result produced after using the generated Overpass Query Hypothesis. An error means that you should definitely improve on the Hypothesis. A normal result could mean that the Overpass Query is good enough, if the output fits to the asked query:
\vspace{-0.1cm}
\begin{lstlisting}
No Results found.
\end{lstlisting}

Improve on the Overpass Query or keep it if it is good enough:\\
\textless \textgreater
\end{mdframed}
\caption{The \textbf{5-shot refine with feedback prompt} used to generate Overpass queries from natural language input. In this case, the execution of the hypothesis did not return any results.}
\label{prompt:5shot_refine_feedback}
\end{figure*}

%% file: tables/refine_feedback_examples.tex
\begin{figure*}[ht]

\begin{mdframed}

\small

\begin{description}
\item[Question:] Cities and towns that are between 100 and 300 km away from Moscow
\item[Hypothesis:]\hfill
\begin{lstlisting}[breaklines, mathescape=true] 
[out:json][timeout:25];(node["place"="city"](around:300000,{{geocodeCoords:"Moscow"}});
node["place"="town"](around:300000,{{geocodeCoords:"Moscow"}}); )$\textbf{; ->}$.all;
(.all; - node["place"~"city|town"](around:100000,{{geocodeCoords:"Moscow"}}););out center;
\end{lstlisting}
\item[Feedback:]
\hfill
\begin{lstlisting}[breaklines, mathescape=true]
Error: line 1: parse error: Unknown type "->"

\end{lstlisting}
\item[New Hypothesis:]\hfill
\begin{lstlisting}[breaklines, mathescape=true]
[out:json][timeout:25]; ( node["place"="city"](around:300000,{{geocodeCoords:"Moscow"}});
node["place"="town"](around:300000,{{geocodeCoords:"Moscow"}}); ) ->.all; 
( .all; - node["place"~"city|town"](around:100000,{{geocodeCoords:"Moscow"}}););out center;

\end{lstlisting}
\end{description}
\hrule

\begin{description}
\item[Question:] All subway stations in the current view
\item[Hypothesis:]\hfill
\begin{lstlisting}[breaklines, mathescape=true] 
[out:json][timeout:25];(
node["railway"="subway_entrance"]({{bbox}});
way["railway"="subway_entrance"]({{bbox}});
relation["railway"="subway_entrance"]({{bbox}});
);out;>;out skel qt;
\end{lstlisting}
\item[Feedback:] \hfill
\begin{lstlisting}[breaklines, mathescape=true] 
{'type': 'node', 'id': 486093610, 'lat': 31.3242632, 'lon': 121.5273796, 'tags': {'railway': 'subway_entrance', 'ref': '3'}}
\end{lstlisting}
\item[New Hypothesis:]\hfill
\begin{lstlisting}[breaklines, mathescape=true]
[out:json][timeout:25];(
node["railway"="$\textbf{station}$"][$\textbf{"subway"="yes"}$]({{bbox}});
way["railway"="$\textbf{station}$"][$\textbf{"subway"="yes"}$]({{bbox}});
relation["railway"="$\textbf{station}$"][$\textbf{"subway"="yes"}$]({{bbox}});
);out;>;out skel qt;
\end{lstlisting}
\end{description}
\hrule
\begin{description}
\item[Question:] pharmacy named Apteka in Wroc\l{}aw
\item[Hypothesis:]\hfill
\begin{lstlisting}[breaklines, mathescape=true] 
[out:json][timeout:120];{{geocodeArea:"Wroc$\text{\l{}}$aw"}}->.searchArea;(
node["amenity"="pharmacy"]["name"="Apteka"](area.searchArea);
way["amenity"="pharmacy"]["name"="Apteka"](area.searchArea);
relation["amenity"="pharmacy"]["name"="Apteka"](area.searchArea);
);out;>;out meta qt;
\end{lstlisting}
\item[Feedback:] No Results found.
% \item [GPT-4 additional commentary:]\hfill
% \begin{lstlisting}[breaklines, mathescape=true]  
% However, the query could still be improved for more robustness. By slightly altering the text search, we can make it case-insensitive and allow for a partial match, which can help in retrieving results where the name may not be exactly \"Apteka\".
% \end{lstlisting}
\item[New Hypothesis:]\hfill
\begin{lstlisting}[breaklines, mathescape=true]
[out:json][timeout:120];{{geocodeArea:"Wroc$\text{\l{}}$aw"}}->.searchArea;(
node["amenity"="pharmacy"]$\textbf{["name"\textasciitilde "Apteka", i]}$(area.searchArea);
way["amenity"="pharmacy"]$\textbf{["name"\textasciitilde "Apteka", i]}$(area.searchArea);
relation["amenity"="pharmacy"]$\textbf{["name"\textasciitilde "Apteka", i]}$(area.searchArea);
);out;>;out meta qt;
\end{lstlisting}
\end{description}
\end{mdframed}

\caption{Examples where the feedback from query execution led to an improved version of the query. \textbf{Top:} The execution of the query raises a syntax error. The symbol "->" is not allowed to occur immediately after a semicolon. This is fixed in the query generated after feedback. \textbf{Middle:} The execution of the query returns results that we append as feedback. However, the result contains the tag subway\_entrance which does not align with the requested subway stations. The refined query then asks for stations instead of subway\_entrances. \textbf{Bottom:} The execution of the hypothesis does not return any results. Given this information, the refined query uses a case-insensitive and partial matching search instead of a strict match.}
\label{tables:refine_examples}
\end{figure*}